\documentclass{article}

 \usepackage[final]{score_workshop}

\usepackage[utf8]{inputenc} 
\usepackage[T1]{fontenc}    
\usepackage{hyperref}       
\usepackage{url}            
\usepackage{booktabs}       
\usepackage{amsfonts}       
\usepackage{nicefrac}       
\usepackage{microtype}      
\usepackage{xcolor}         
\usepackage{amsmath,graphicx,caption,mathtools,amssymb,amsthm}
\usepackage{subcaption}
\usepackage{comment}
\usepackage{wrapfig}
\title{Progressive Deblurring of Diffusion Models for Coarse-to-Fine Image Synthesis}

\newtheorem{proposition}{\bf{Proposition}}

\newcommand{\B}{\mathbf{B}}

\newcommand{\U}{\mathbf{U}}

\newcommand{\x}{\mathbf{x}}
\newcommand{\f}{\mathbf{f}}

\newcommand{\w}{\mathbf{w}}
\newcommand{\z}{\mathbf{z}}

\newcommand{\Ab}{{\mathbf A}}
\newcommand{\Bb}{{\mathbf B}}
\newcommand{\Cb}{{\mathbf{C}}}
\newcommand{\Db}{{\mathbf D}}

\newcommand{\Gb}{{\mathbf G}}
\newcommand{\Hb}{{\mathbf H}}
\newcommand{\Ib}{{\mathbf I}}

\newcommand{\Pb}{{\mathbf P}}

\newcommand{\Ub}{{\mathbf U}}

\newcommand{\Wb}{{\mathbf W}}

\newcommand{\Nc}{\mathcal{N}}

\newcommand{\beq}{\begin{equation}}
\newcommand{\eeq}{\end{equation}}
\newcommand{\beqa}{\begin{eqnarray}}
\newcommand{\eeqa}{\end{eqnarray}}

\author{%
  Sangyun Lee \\
  Soongsil University\\
  \texttt{ml.swlee@gmail.com} \\
   \\
   \\
  \And
  Hyungjin Chung \\
  Dept. of Bio and Brain Engineering \\
  KAIST, Korea\\
  \texttt{hj.chung@kaist.ac.kr} \\
  \\
  \AND
  Jaehyeon Kim \\
  Kakao Enterprise \\
  \texttt{jay.xyz@kakaoenterprise.com} \\
  \\
  \And
  Jong Chul Ye \\
  Kim Jaechul Graduate School of AI \\
KAIST,   Korea\\
  \texttt{jong.ye@kaist.ac.kr} \\
}

\begin{document}

\maketitle

\begin{center}
    \centering

    \includegraphics[width=1\linewidth]{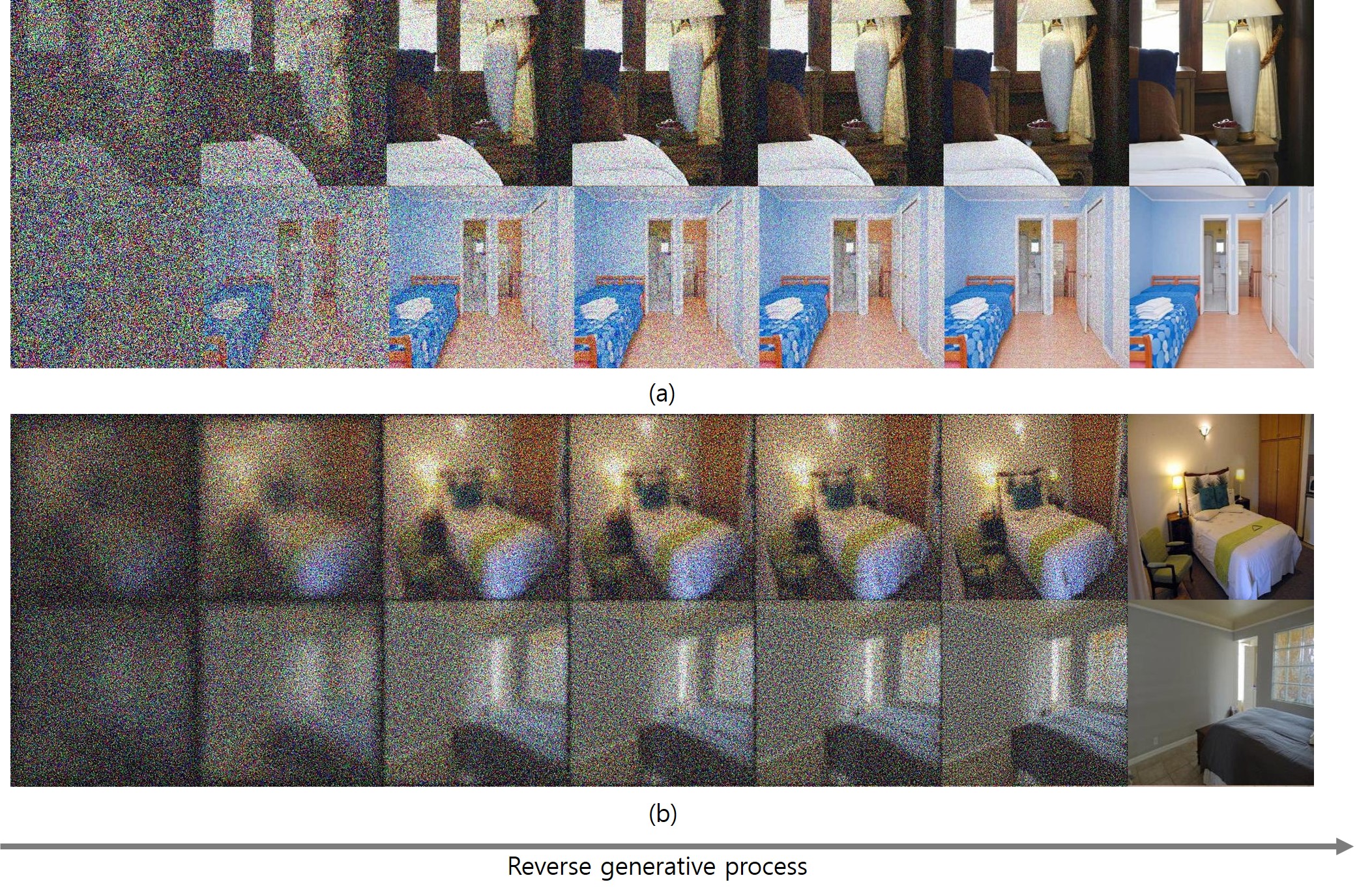}
    \captionof{figure}{Reverse generative processes of two different diffusion models. (a) Previous diffusion models generate images by gradually strengthening signals. (b) The proposed method synthesizes images through progressive deblurring in a coarse-to-fine manner.}

    \label{fig:main}
\end{center}

\begin{abstract}
  Recently, diffusion models have shown remarkable results in image synthesis
by gradually removing noise and amplifying signals. 
Although the simple generative process surprisingly works well, is this the best way to generate image data?
For instance, despite the fact that human perception is more sensitive to the low-frequencies of an image, diffusion models themselves do not consider any relative importance of each frequency component.
Therefore, to incorporate the inductive bias for image data, we propose a novel generative process that synthesizes images in a coarse-to-fine manner.
First, we generalize the standard diffusion models by enabling diffusion in a rotated coordinate system with different velocities for each component of the vector.
We further propose a \textit{blur diffusion} as a special case, where each frequency component of an image is diffused at different speeds.
Specifically, the proposed blur diffusion consists of a forward process that blurs an image and adds noise gradually, after which  a corresponding reverse process deblurs an image and removes noise progressively.
Experiments show that proposed model outperforms the previous method in FID on LSUN bedroom and church datasets.
\end{abstract}

\section{Introduction}

After the initial development by \citet{sohl2015deep}, diffusion models have been rapidly improved~\citep{ho2020denoising,dhariwal2021diffusion,song2019generative,song2020score} to the point that they achieve superior results than GANs in both fidelity and diversity in image synthesis.
Since these models  offer better mode coverage, 
they are widely used in various tasks such as image generation~\citep{dhariwal2021diffusion}, super resolution~\citep{li2022srdiff,saharia2021image},  text-conditional image generation~\citep{nichol2021glide,ramesh2022hierarchical}, video generation~\citep{ho2022video}, etc.

Since a forward process of diffusion models attenuates signals by adding noise progressively, a reverse process generates data by gradually removing noise and amplifying signals.
Although this formulation gives a great simplicity (e.g., no need to deal with the covariance matrix) and surprisingly works well, it may not be the best way to generate image data.
For instance, despite the fact that human perception is more sensitive to the low-frequencies of an image, diffusion models themselves do not consider any relative importance of each frequency component.

To incorporate the inductive bias for image data, several methods have been suggested to focus on  coarse patterns of an image to improve the perceptual quality of generated samples.
For instance, diffusion models are usually trained on a re-weighted variational lower bound~\citep{ho2020denoising}, which emphasizes the global consistency and coarse level pattern of images and gives less focus on the imperceptible details~\citep{kingma2021variational}.
The performance of diffusion models can be greatly improved by adopting a coarse-to-fine strategy, where a low-resolution image is generated first and then upsampled by separate diffusion upsamplers~\citep{ho2022cascaded,dhariwal2021diffusion,nichol2021improved,nichol2021glide,ramesh2022hierarchical}.
By explicitly partitioning the generative process into the stage of generating coarse structure and the stages of adding details, these models are capable of producing convincing images, especially at high-resolution.

However, dividing into the predetermined number of stages is somewhat arbitrary and requires learning separate upsampler for each stage.
In this paper, we propose a novel generative process that synthesizes images in a coarse-to-fine manner.
Our model does not require any upsamplers or separate stages.
Instead, we generalize the standard diffusion models by enabling diffusion in a rotated coordinate system with different velocities for each component of the vector.
We further propose a \textit{blur diffusion} as a special case of it, where each frequency component of an image is diffused at different speeds.
In particular, our blur diffusion consists of a forward process that blurs an image while adding noise gradually, and a corresponding reverse process that deblurs an image while removing noise progressively.
Experiments show that the proposed model outperforms the previous method in FID on LSUN bedroom and church datasets (64$\times$64).

\section{Blur diffusion}
\label{blur}
Coarse-to-fine generation in image synthesis is a successful strategy for both GANs~\citep{karras2017progressive,karras2019style} and diffusion models~\citep{ho2022cascaded,dhariwal2021diffusion,nichol2021improved,nichol2021glide,ramesh2022hierarchical}.
The most intuitive way to enable the strategy without separate stages is to define a gradual blurring forward process and reverse it.
This can be seen as diffusion in a rotated coordinate system with different velocities for each component of the vector.
We first introduce a generalized diffusion process (Sec.~\ref{sec:gen_diffusion}) and propose the blur diffusion as a special case of it (Sec.~\ref{sec:blur_diffusion}).


\subsection{Generalized diffusion}
\label{sec:gen_diffusion}
Before introducing our contributions, we refer the readers to Appendix~\ref{background} for a background on diffusion models.
For each training data $\mathbf x_0 \sim q_0(\mathbf x),$ a forward process of the variance preserving diffusion models~\citep{song2020score} is defined from the following Markov chain:
\begin{equation}
    \mathbf x_i = \sqrt{1-\beta_i}\mathbf x_{i-1} + \sqrt{\beta_i}\mathbf z_{i}, \ \ i=1,...,N
    \label{Equation:forward-previous}
\end{equation}
where $z_i\sim \mathcal N(0,\mathbf I)$, and $\{\beta_i\}_{i=1}^N$ is a pre-defined noise schedule.
As one can see, a standard diffusion process is defined in the image space directly, assuming the independence between each pixels. Our aim is to generalize this process in a rotated coordinate system.
For this,  we define an orthogonal matrix $\U$, and subsequently some vector rotated by the matrix as $\bar\x := \U^T\x$.
With slight abuse of notation, we define the fractional powers of a positive semi-definite matrix $\Pb^p$ as taking the powers of each eigenvalue, i.e, $\Pb^{p} = (\Ub \mathbf{\Lambda} \Ub^T)^{p}=\Ub \mathbf{\Lambda}' \Ub^T$, where $[\mathbf{\Lambda}']_{ii}=[\mathbf{\Lambda}]_{ii}^p$.

Then, we define a generalized forward diffusion process with the following Markov chain:
\begin{align}
\label{eq:gen_diff}
    q(\bar\x_i |\bar\x_{i-1}) = \Nc(\bar\x_i; (\Ib - \Bb_i)^{\frac{1}{2}}\bar\x_{i-1}, \Bb_i\Ib),
\end{align}
where $\Bb_i$ is a diagonal matrix that defines the noise schedule of the process. Note that \eqref{eq:gen_diff} is a generalized version of the standard diffusion, as standard diffusion is retrieved when we set $\Ub = \Ib$, and $\Bb_i = \beta_i \Ib$. In other words, we are introducing more flexibility into the design space of diffusion models by enabling 1) diffusion in the rotated coordinate, where the dependency between pixels can be imposed, 2) diffusion with different velocities for each component of the vector.

Due to the properties of diagonal matrices, we arrive at an analytically tractable conditional distribution
\begin{align}
\label{eq:qxi_given_x0}
    q(\bar\x_{i}|\bar\x_0) = \Nc(\bar\x_i; \bar\Ab_i^{\frac{1}{2}}\bar\x_0, (\Ib - \bar\Ab_i)),
\end{align}
where we have defined $\Ab_i := \Ib - \Bb_i$, and $\bar\Ab_i := \prod_{j=1}^i \Ab_j$, analagous to~\citep{ho2020denoising}. 
Eq.~\eqref{eq:qxi_given_x0} allows one to directly calculate $\x_i$ using $\x_0$:
\begin{equation}
    \x_i = \Ub\bar\Ab_i^{\frac{1}{2}}\Ub^T\x_0 + \Ub(\Ib - \bar\Ab_i)^{\frac{1}{2}} \Ub^T \mathbf{\epsilon},
    \label{eq:rot_xi_x0}
\end{equation}
where $\mathbf{\epsilon} \sim \mathcal N(0,\mathbf I)$.
Tractability of Eq.~\eqref{eq:rot_xi_x0} in turn means that we can efficiently train these models with denoising score matching~\citep{vincent2011connection} as in prior studies.

\subsection{Blur diffusion}
\label{sec:blur_diffusion}

While the choice of the rotation matrix $\Ub$ and the noise schedule $\B_i$ are flexible, here we propose an especially effective choice that can be characterized as blurring diffusion for the forward process. For simplicity and ease of computation, we utilize Gaussian blur with symmetric kernels that are separable, with a pre-defined variance of $\sigma^2$. 
Since Gaussian blur is a linear operation, it can be approximated as a matrix multiplication using a circular symmetric matrix $\mathbf W$.
With some monotonically increasing function $f(i)$ that determines a blur schedule and $\Wb_i = \Wb^{f(i)}$, we define a blurring diffusion process as follows:
\begin{align}
\label{eq:blur_diff}
    q(\x_i|\x_{i-1}) = \Nc(\x_i; \sqrt{1-\beta_i}\Wb_i\x_{i-1}, \Cb_i),
\end{align}
where we set $\Cb_i = \Ib - (1-\beta_i)\Wb_i^2$ to ensure the process preserves unit variance.
Eq.~\eqref{eq:blur_diff} can also be written as follows:
\begin{equation}
\label{eq:blur_diff_discretize}
    \mathbf x_i = \mathbf x_{i-1} -\mathbf H(\mathbf x_{i-1}, i-1) +  \Cb_i^{\frac{1}{2}}\z_i,
\end{equation}
where $\mathbf H(\mathbf x_{i}, i) = \mathbf x_{i} - \sqrt{1-\beta_{i+1}}\Wb_{i+1}\x_i$ is an unnormalized Gaussian high-pass filter.
Unlike the standard forward diffusion process where the signal is attenuated holistically (see eq.~\eqref{Equation:forward-previous2}), our forward process destroys high frequencies much faster.
In order to match the definition of the generalized diffusion, we propose to factor the symmetric matrix $\Wb$ by eigenvalue-decomposition $\Wb = \tilde\Ub\Db\tilde\Ub^T$ and subsequently $\Wb_i = \tilde\Ub\Db^{f(i)}\tilde\Ub^T$. We employ the memory-efficicent eigen-decompisition by \citet{kawar2022denoising} (see Appendix D of DDRM).
This leads us to the following proposition

\begin{proposition}
\label{prop:blur}
Let $\Bb_i = \Ib - (1-\beta_i) \Db^{2f(i)}$ and $\Ub = \tilde\Ub$.
Then, \eqref{eq:gen_diff} is equivalent to \eqref{eq:blur_diff}.
\end{proposition}
Proof is deferred to Appendix~\ref{sec:proof}.
Due to Proposition~\ref{prop:blur}, we can efficiently train the model using the denoising score matching objective:
\begin{equation}
    \mathcal L = \mathbb E_i\{\lambda(i) \mathbb E_{q_0(\mathbf x)} \mathbb E_{q(\mathbf x_i | \mathbf x_0)}[||\mathbf s_{\mathbf \theta}(\mathbf x_i, i) - \underbracket{ (-\mathbf  U(\mathbf I-\bar {\mathbf A_i})^{-1}\mathbf U^T \mathbf \epsilon)}_{=\nabla_{\mathbf x_i} \log  { q_i(\mathbf x_i | \mathbf x_0)})}   ||_2^2]\}.
    \label{Equation:dsm_blur}
\end{equation}
When we parameterize $\mathbf{s}_\theta(\x_i,i)$ as
\begin{equation}
    \mathbf{s}_\theta(\x_i,i) = -\Ub(\Ib - \bar \Ab_i)^{-1}\Ub^T \boldsymbol{\epsilon}_\theta(\x_i,i),
\end{equation}
Eq.~\eqref{Equation:dsm_blur} is simplified to:
\begin{equation}
    \mathbb E_i\{\lambda(i) \mathbb E_{q_0(\mathbf x)} \mathbb E_{q(\mathbf x_i | \mathbf x_0)}[|| \mathbf  U(\mathbf I-\bar {\mathbf A_i})^{-1}\mathbf U^T  ( \boldsymbol{\epsilon}_\theta(\mathbf x_i, i) - \boldsymbol{\epsilon})
    ||_2^2]\}.
    \label{Equation:L_eps}
\end{equation}
In practice, we found it beneficial to sample quality to use the following variant of Eq.~\eqref{Equation:L_eps}:
\begin{equation}
    \mathcal L_{\epsilon} = \mathbb E_i\{\lambda(i) \mathbb E_{q_0(\mathbf x)} \mathbb E_{q(\mathbf x_i | \mathbf x_0)}[||\boldsymbol{\epsilon}_\theta(\mathbf x_i, i) -  \boldsymbol{\epsilon}
    ||_2^2]\},
    \label{Equation:L_eps_simple}
\end{equation}
which resembles a re-weighted VLB~\citep{ho2020denoising}.

\paragraph{Reverse deblurring process}
After training, we can sample images using a reverse diffusion sampler:
\begin{equation}
    \mathbf x_{i-1}=\mathbf x_i + \mathbf H(\mathbf x_i, i) + \mathbf U\mathbf B_{i+1}\mathbf U^T\mathbf s_\theta(\mathbf x_i, i) + \mathbf U\mathbf B_{i+1}^{\frac{1}{2}}\mathbf U^T\mathbf z_{i}
\end{equation}
or equivalently,
\begin{equation}
    \mathbf x_{i-1}=\mathbf x_i + \mathbf H(\mathbf x_i, i) - \mathbf U\mathbf B_{i+1} (\Ib - \bar \Ab_i) \mathbf U^T \boldsymbol{\epsilon}_\theta(\mathbf x_i, i) + \mathbf U\mathbf B_{i+1}^{\frac{1}{2}}\mathbf U^T\mathbf z_{i},
\end{equation}
which is analogous to unsharp masking~\citep{szeliski2010computer} followed by the denoising term to remove amplified noise.
Through the process, our model generates an image in a coarse-to-fine manner by progressive deblurring followed by denoising (see Figure~\ref{fig:main}).

\begin{table}[!hbt]
\centering
\begin{tabular}{@{}llll@{}}
\toprule
             &         & \multicolumn{2}{c}{FID-10K}           \\
$f(N)$         & $f$\_type & \multicolumn{1}{l|}{bedroom} & church \\ \midrule
0 (w/o blur) & N/A     & \multicolumn{1}{l|}{9.24}    & 6.04   \\
0.6           & log     & \multicolumn{1}{l|}{73.23}   &        \\
0.14           & quartic & \multicolumn{1}{l|}{\textbf{7.86}}    & \textbf{5.89}   \\ \bottomrule
\end{tabular}
\caption{FID-10K results on LSUN bedroom and church-outdoor datasets (64$\times$64). We fix $f(0)$ to 0.}
\label{Table:ablation}
\end{table}

\section{Comparison with standard diffusion models}
We note that standard diffusion models are the special case of our model when $f(i)=0$.
Table~\ref{Table:ablation} demonstrates that our model outperforms the standard diffusion model when $f$\_type is quartic with $f(N)=0.14$.
We provide a functional form for each $f$\_type in Appendix.
We compute FID using only 10K samples, and this is acceptable as we measure the relative differences within the same framework.
See Appendix~\ref{sec:experiment} for additional experiments.

\section{Conclusion}
In this paper, we generalized the previous diffusion models and provide an effective way to impose the inductive bias on diffusion models.
We further proposed the blur diffusion as a special case.
Blur diffusion generates images in a coarse-to-fine manner by progressive deblurring followed by denoising.
Experiments showed that our model can synthesize more perceptually compelling samples than previous methods.
We look forward to scaling up and applying the model to various applications.

\section*{Acknowledgement}
This work was supported by the National Research Foundation of Korea under Grant NRF-2020R1A2B5B03001980, and by the KAIST Key Research Institute (Interdisciplinary Research Group) Project.

\bibliography{reference}



\newpage
\appendix
\begin{center}
      {\bf APPENDIX}
\end{center}

\section{Background}
\label{background}
In this section, we briefly overview the variance preserving diffusion models~\citep{song2020score}.
For each training data $\mathbf x_0 \sim q_0(\mathbf x),$ a forward process is defined from the following Markov chain:
\begin{equation}
    \mathbf x_i = \sqrt{1-\beta_i}\mathbf x_{i-1} + \sqrt{\beta_i}\mathbf z_{i}, \ \ i=1,...,N
    \label{Equation:forward-previous}
\end{equation}
where $z_i\sim \mathcal N(0,\mathbf I)$ and $\{\beta_i\}_{i=1}^N$ is a pre-defined noise schedule.
Another way to see Eq.~\eqref{Equation:forward-previous} is:
\begin{equation}
    \mathbf x_i = \mathbf x_{i-1} - (1 - \sqrt{1-\beta_i})\mathbf x_{i-1} + \sqrt{\beta_i}\mathbf z_{i},
    \label{Equation:forward-previous2}
\end{equation}
which means that each step of the forward process consists of attenuating signals holistically and adding Gaussian noise.
It is noteworthy that $q(\mathbf x_i|\mathbf x_0)$ is also Gaussian distribution and can be written in a closed form, allowing efficient training. 
Specifically, using the notation $\alpha_i = 1-\beta_i$ and $\bar \alpha_i = \prod_{j=1}^i \alpha_i$, we have
\begin{equation}
    q(\mathbf x_i|\mathbf x_0) = \mathcal N(\mathbf x_i;\sqrt {\bar \alpha_i}\mathbf x_0, (1-\bar \alpha_i)^2\mathbf I).
\end{equation}
To generate clean images, we need to invert the noising process using sampling methods, which requires estimating the time-conditional score function $\nabla_{\mathbf x} \log  {q_i(\mathbf x)}$ where $q_i(\mathbf x)=\int q(\mathbf x_i|\mathbf x_0)q_0(\mathbf x_0)d\mathbf x_0$.
One way to estimate the score function is to minimize the denoising score matching objective~\citep{vincent2011connection}:
\begin{equation}
    \mathbf \theta^* = \arg \min_{\mathbf \theta} \mathbb E_i\{\lambda(i) \mathbb E_{q_0(\mathbf x)} \mathbb E_{q(\mathbf x_i | \mathbf x_0)}[||\mathbf s_{\mathbf \theta}(\mathbf x_i, i) - \nabla_{\mathbf x} \log  { q_i(\mathbf x_i | \mathbf x_0)}||_2^2]\},
\end{equation}
where $\lambda (i)$ is a non-negative weighting function.
When $\mathbf s_\theta(\mathbf x_i,i)$ is a reasonable predictor of the score function, sampling can be done using, for instance, a reverse diffusion sampler~\cite{song2020score}:
\begin{equation}\label{eq:sample}
    \mathbf x_{i-1}=\mathbf x_i-(\sqrt{1-\beta_{i+1}}-1)\mathbf x_i + \beta_{i+1}\mathbf s_\theta(\mathbf x_i, i) + \sqrt{\beta_{i+1}}\mathbf z_{i}.
\end{equation}
Another popular sampling method is ancestral sampling~\citep{ho2020denoising}, which is a different discretization of the same reverse-time stochastic differential equation (SDE)~\citep{song2020score}. Since the diffusion sampler \eqref{eq:sample} can be derived in a conceptually simple manner for an arbitrary SDE, we extensively use it in this paper.

\section{Experiments}
\label{sec:experiment}

\begin{figure}
  \centering
    \begin{tabular}{cc}
    \includegraphics[width=0.5\linewidth]{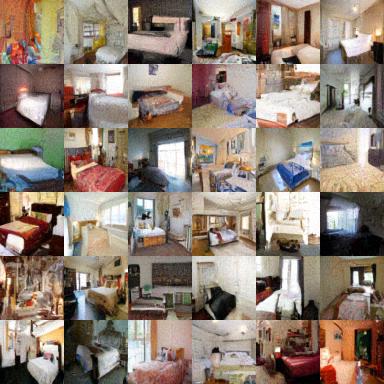}  & \includegraphics[width=0.5\linewidth]{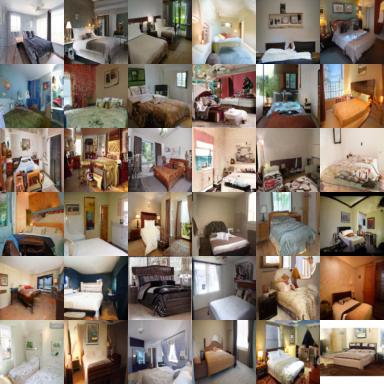}

    \end{tabular}
    \caption{Results on LSUN-bedroom 64 $\times$ 64. f\_type : log (\textit{left}), f\_type : quartic (\textit{right}).}
    \label{fig:bedroom}
\end{figure}

\begin{figure}
  \begin{center}
  \begin{tabular}{cc}
    \includegraphics[width=0.5\linewidth]{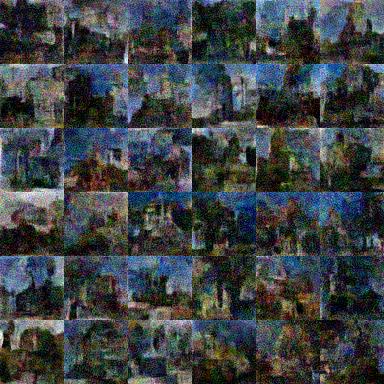}  & \includegraphics[width=0.5\linewidth]{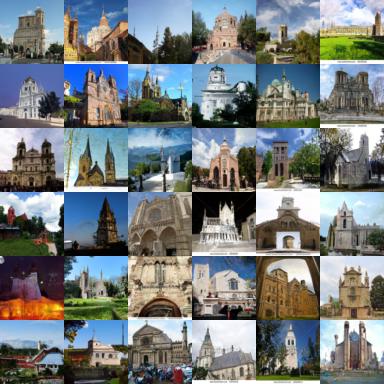}

    \end{tabular}
    
  \end{center}
  \caption{Comparison of generated images with different generation strategies. Left: fine-to-coarse, right: coarse-to-fine.}
  \label{fig:fine-to-coarse}
\end{figure}

\paragraph{Experiment details}
We set $N=1000$ and sample using $N$ steps for all experiments.
All models are trained on a single V100 with a batch size of 16.
We train models for 450K and 600K iterations on LSUN bedroom and church datasets, respectively.
We set the learning rate to $5e-5$, EMA decay factor to 0.9999, $\sigma$ to 0.4, and $\lambda(i)$ to 1.
For pre-processing, we resize images to $64\times 64$ without cropping.
We do not use any dropout in our experiments.
We provide detailed model configuration and computational requirements in Appendix.

\paragraph{Comparison of different sampling strategy}
We conduct an experiment where we compare our coarse-to-fine approach with fine-to-coarse strategy, for which we replace the diagonal matrix $\Db$ with $\Ib-\Db$.
As shown in Fig.~\ref{fig:fine-to-coarse}, the fine-to-coarse strategy results in significant artifacts in the generated images, indicating that it is crucial to impose the appropriate inductive bias on diffusion models.

\paragraph{A form of $f$}

\begin{wrapfigure}[14]{r}{0.3\textwidth}
    \centering
    \includegraphics[width=1\linewidth]{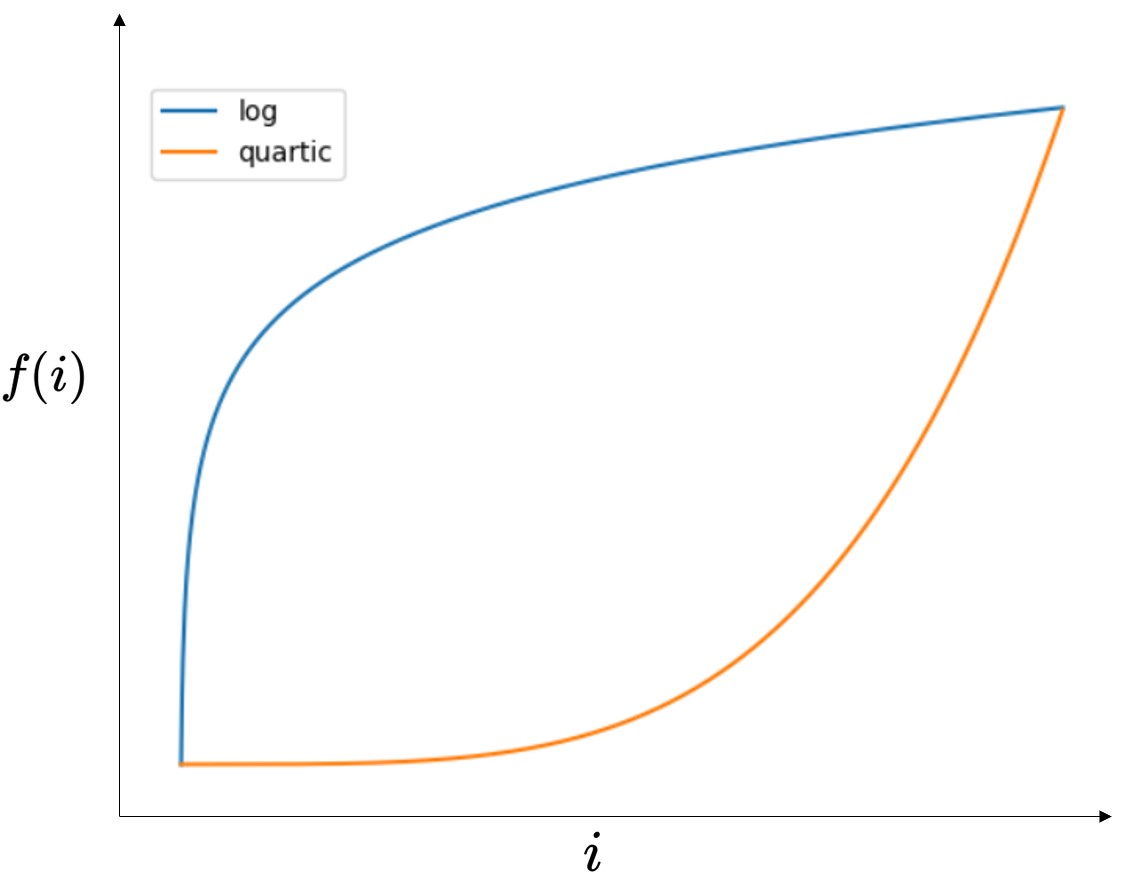}

    \caption{Different functional forms of blur schedule $f(i)$ we experimented with.}

    \label{fig:ftype}
\end{wrapfigure}
Figure~\ref{fig:ftype} shows the several functional forms of blur schedule we used. Given the fixed $f(N)$, we set $f(i) = \frac{f(N)}{N^4}i^4$ and $f(i) = \frac{f(N)}{\log N} \log i$ for the quartic and log settings, respectively.
As shown in Figure~\ref{fig:bedroom} and Table~\ref{Table:ablation}, increasing the blur strength too early leads to inferior sample qualities: the model fails to generate reliable high frequencies.

\section{Discussion}
\paragraph{Perceptual quality}
While diffusion models can be trained in a perceptual quality-oriented way using hand-crafted weighting functions, our method provides a more explicit way to focus on the coarse pattern of images: to train a score estimator on the blurred images.
Moreover, our model emphasizes the coarse patterns during the sampling process as the score function points in the steepest direction to the high-density region of the low-frequencies, especially when $i$ is close to $N$.
Since we did not sweep over blur schedules other than what we reported, it would be a valuable future direction to further find the optimal noise and blur schedule for our model.
\paragraph{Different basis}
Although we mainly discuss the blur diffusion as a special case, our generalized diffusion is broadly applicable to the arbitrary coordinate depending on the choice of the orthonormal basis $\Ub$.
For instance, one can perform diffusion in different frequency domains of such as Fourier transform or discrete cosine transform.
Moreover, our approach is not restricted to the image data and can be used for different data modalities as we provide a general method for imposing the inductive bias on diffusion models.

\section{Related works}
\label{sec:related}
Several approaches have been proposed to find a better space for diffusion models.
Recently, \citet{vahdat2021score} and \citet{rombach2022high} proposed to train the diffusion model in the learned latent space.
Unlike our method, these approaches require the training of an autoencoder and have no control over the learned space, which is necessary to impose the inductive bias for a certain data modality of interest.
\citet{jing2022subspace} utilize the orthogonal projection to destroy the component orthogonal to the data manifold faster.
Unlike our work, they focus on reducing the costs for sampling, and the method requires the predetermined time steps in which the projection is performed.

\citet{rissanen2022generative} concurrently proposed a deblurring generative process by reversing the heat equation.
Indeed, their method is a special case of our generalized diffusion, in which the columns of $\Ub$ are cosine basis.
Although solving heat equation does not involve noise, they empirically found that a small amount of noise (with the variance of 0.1) in the forward process as well as in the generative process is crucial for the sample quality.
The noise strength for both processes is chosen by trial and error.
In contrast, our approach naturally involves noise as we interpret the proposed diffusion process from the SDE perspective.
Therefore, once the noise schedule of the forward process is determined, the reverse-time noise strength is rigorously derived from the forward process, while \citet{rissanen2022generative} swept over the reverse-time noise strength $\delta$.
Finally, their novel iterative method does not show any improvements in performance yet, while our method demonstrates the improved performance as a generalization of standard diffusion models.

\section{Proof}
\label{sec:proof}
In this section, we provide a proof for Proposition~\ref{prop:blur}.
With Eq.~\eqref{eq:gen_diff}, $\bar\x_i$ is represented as follows:

\begin{equation}
\bar\x_i = \sqrt{1-\beta_i}\Db^{f(i)}\bar\x_{i-1} + (\Ib - (1-\beta_i)\Db^{2f(i)})^{\frac{1}{2}}\z_i    
\end{equation}
Using the definition of $\bar\x_i$, we have
\begin{gather}
    \x_i = \sqrt{1-\beta_i}\tilde\Ub \Db^{f(i)}\tilde \Ub^T \x_{i-1} + \tilde \Ub(\Ib - (1-\beta_i)\Db^{2f(i)})^{\frac{1}{2}}\tilde \Ub^T\bar \z_i \\
    = \sqrt{1-\beta_i}\tilde\Ub \Db^{f(i)}\tilde \Ub^T \x_{i-1} + (\tilde \Ub(\Ib - (1-\beta_i)\Db^{2f(i)})\tilde \Ub^T)^{\frac{1}{2}}\bar \z_i \\
    = \sqrt{1-\beta_i}\Wb_i\x_{i-1} + \Cb_i^{\frac{1}{2}}\bar \z_i,
\end{gather}
where $\bar \z_i \sim \mathcal N(0,\Ib)$.
Note that $\Cb_i = \Ib - (1-\beta_i)\tilde \Ub \Db^{2f(i)}\tilde \Ub^T = \tilde \Ub (\Ib - (1-\beta_i)\Db^{2f(i)})\tilde \Ub^T$.

\section{Derivation of reverse diffusion sampler}
It was shown in~\cite{song2020score} that in the continuous time limit, diffusion models can be viewed as realizations of SDEs. In such view, generative inference can be regarded as discretized solutions to the reverse SDEs. We follow the discretization rules from~\cite{song2020score} henceforth to derive our reverse diffusion sampler. Specifically, for a vector-valued function $\f_i(\cdot)$ and matrix $\Gb_i$, consider the following stochastic difference equation
\begin{equation}
 \x_{i+1} = \x_{i} + \f_i(\x_i,i) + \Gb_i \z_{i},
 \label{eq:difference}
\end{equation}
where $\z_{i} \sim \mathcal N(\mathbf{0}, \Ib)$.
With $\bar \f(\x_i,i) = N \cdot \f_i(\x_i,i)$ and $\bar \Gb_i = \sqrt{N} \cdot \Gb_i$, Eq.~\eqref{eq:difference} can be written as
\begin{equation}
    \x_{i+1} - \x_i = \bar \f(\x_i, i) \Delta t + \bar \Gb_i \sqrt{\Delta t} \z_{i},
\end{equation}
where $\Delta t = \frac{1}{N}$.
Let $\x(t) = \x_i, \Gb(t) = \bar \Gb_i,$ and $\f(\x(t), t) = \bar \f(\x_i,i)$ for $t = \frac{1}{N}$.
In the limit of $N \rightarrow \infty$, $\{\x_i\}_i$ becomes a continuous-time stochastic process $\x(t)$ governed by following SDE:
\begin{equation}
    d\x = \f(\x(t),t)dt + \Gb(t)d\bar \w.
    \label{eq:appendix-sde}
\end{equation}
Anderson's theorem~\citep{anderson1982reverse} leads us to the following reverse-time SDE of Eq.~\eqref{eq:appendix-sde}:
\begin{equation}
    d\x = [\f(\x,t) - \Gb(t) \Gb(t)^T \nabla_\x \log p_t(\x)]dt + \Gb(t)d\w.
    \label{eq:appendix-reverseSDE}
\end{equation}
For sampling, we further discretize Eq.~\eqref{eq:appendix-reverseSDE} as follows:
\begin{equation}
    \x_{i-1} = \x_i - \f_i(\x_i, i) + \Gb_i \Gb_i^T \mathbf{s}_\theta(\x_i, i) + \Gb_i \z_i,
\end{equation}
and this is called the reverse diffusion sampler.
For blur diffusion, we set $\f_i(\x_i,i) = -\Hb(\x_i,i)$ and $\Gb_i = \Cb_{i+1}^{\frac{1}{2}} = \tilde \Ub \Bb_{i+1}^{\frac{1}{2}} \tilde \Ub^T$ from Eq.~\eqref{eq:blur_diff_discretize}, and thus we have:
\begin{equation}
    \x_{i-1} = \x_i + \Hb(\x_i, i) + \tilde \Ub \Bb_{i+1} \tilde \Ub^T \mathbf{s}_\theta(\x_i, i) + \tilde \Ub \Bb_{i+1}^{\frac{1}{2}} \tilde \Ub^T \z_i.
\end{equation}

\section{Architecture configuration}
We utilize the UNet architecture of \citet{dhariwal2021diffusion}.
A detailed architecture configuration is as follows:

\begin{itemize}
    \item Diffusion steps: 1000
    \item Noise schedule: linear
    \item Model size: 121M
    \item Channels: 128
    \item Depth: 3
    \item Channels multiple: 1,2,3,4
    \item Heads: 4
    \item Attention resolution: 4,8
    \item BigGAN up/downsampling: False
    \item Dropout: 0
    \item Batch size: 16
    \item Iterations: 450K for LSUN bedroom, 600K for LSUN church
    \item Learning rate: 1e-5
\end{itemize}
All models are trained on a single V100.

\end{document}